\documentclass[runningheads]{llncs}
\usepackage[T1]{fontenc}
\usepackage{graphicx}
\usepackage{cite}

\usepackage{algorithm}
\usepackage{algpseudocode}
\usepackage{multirow}
\usepackage{amsmath,amssymb,amsfonts}
\usepackage{graphicx}
\usepackage{textcomp}
\usepackage{xcolor}
\usepackage[utf8]{inputenc}
\usepackage{graphicx}
\usepackage{lineno,hyperref}
\usepackage{amssymb}
\usepackage{graphicx}
\usepackage{multirow}
\usepackage{mathtools}

\DeclareMathOperator*{\argmin}{arg\,min}
\begin{document}
\title{An Approach Towards Learning K-means-friendly
Deep Latent Representation}
%
%
\author{Debapriya Roy\inst{1}\orcidID{0000-0001-8657-3130} 
}
\authorrunning{Roy et al.}
%
\institute{Institute of Engineering and Management, Kolkata, India
\email{debapriyakundu1@gmail.com}\\
}
\maketitle              
\begin{abstract}
Clustering is a long-standing problem area in data mining. The centroid-based classical approaches to clustering mainly face difficulty in the case of high dimensional inputs such as images. With the advent of deep neural networks, a common approach to this problem is to map the data to some latent space of comparatively lower dimensions and then do the clustering in that space. Network architectures adopted for this are generally autoencoders that reconstruct a given input in the output. To keep the input in some compact form, the encoder in AE's learns to extract useful features that get decoded at the reconstruction end. A well-known centroid-based clustering algorithm is K-means. In the context of deep feature learning, recent works have empirically shown the importance of learning the representations and the cluster centroids together. However, in this aspect of joint learning, recently a continuous variant of K-means has been proposed; where the softmax function is used in place of argmax to learn the clustering and network parameters jointly using stochastic gradient descent (SGD). However, unlike K-means, where the input space stays constant, here the learning of the centroid is done in parallel to the learning of the latent space for every batch of data. Such batch updates disagree with the concept of classical K-means, where the clustering space remains constant as it is the input space itself. To this end, we propose to alternatively learn a clustering-friendly data representation and K-means based cluster centers. Experiments on some benchmark datasets have shown improvements of our approach over the previous approaches.
\end{abstract}

\section{Introduction}
\label{introduction}
Clustering is a method of finding the inherent pattern in data by segregating it into different groups. Primarily, it is used to partition unlabeled data into groups for extracting meaningful information. It has various applications in recommender systems where user queries are often grouped to give informed product suggestions, resulting in a better purchasing pattern. In information retrieval systems, this is also used for partitioning similar or associated articles into the same clusters to enhance the efficiency and effectiveness of the retrieval process. In addition, clustering has immense applicability in image segmentation, medical imaging, social network analysis, anomaly detection, market segmentation, etc.

The idea of grouping similar items needs some distance metric. As the complex manifold in high-dimensional input space makes the use of Euclidean distance less meaningful, therefore, classical clustering methods like K-means and GMM are less effective there. This brings up the idea of clustering in latent space, a comparatively low-dimensional space. However, learning in such a space is often challenging as we are dealing with an unsupervised problem. Employing autoencoders (AE) in learning the "clustering friendly" latent space has facilitated the development of deep clustering approaches over the last few years~\cite{dkm_fard2020deep, dcn_yang2017towards, idec_guo2017improved}.

Based on the ways to learn an embedding space with latent representations suitable for clustering, the existing approaches to deep clustering can be primarily categorized into three types. The first type of method~\cite{dec_xie2016unsupervised} learns the latent representations first in the pretraining phase, then optimizes the representations for clustering using some clustering loss while learning the clustering parameters. This type of method finetunes the embedding space for clustering without regard to its reconstructability. The very first work in this direction is Deep Embedded Clustering (DEC)~\cite{dec_xie2016unsupervised}. DEC employs a clustering loss to finetune the encoder of the pretrained AE for clustering while learning the cluster centers. Initially, the soft assignments between the embedded points and the cluster centers are computed. The KL divergence-based clustering loss is employed to improve upon this initial soft estimate by learning from high-confidence predictions. This is done by pushing the soft estimates toward the hard estimates. As DEC abandons the decoder and finetunes the encoder using only the clustering loss, this might distort the embedded space, causing a loss of representativeness of the data. But employing the clustering loss only for finetuning could distort the embedded space to the extent of weakening the representativeness of the latent features, which in turn could hurt the clustering performance. 

Considering this, the second type of method~\cite{idec_guo2017improved,dcn_yang2017towards} proposed a joint autoencoder (AE)-based dimensionality reduction (DR) and K-means objective. In IDEC, the DR and cluster center learning are done jointly. Whereas the target distribution is updated every T iterations of DR and cluster center updates. Unlike DEC and IDEC, where the clustering loss is based on KL-divergence, DCN~\cite{dcn_yang2017towards} adopted a clustering loss related to the classical K-means. However, in K-means, the membership values are restricted to a discrete set, causing the joint optimization of clustering and the reconstruction loss to be numerically infeasible. DCN deals with this by jointly optimizing the reconstruction and the clustering loss in alternating stochastic optimization, where gradient update and discrete cluster assignments are done alternatively. In DCN, this is optimized using alternative stochastic optimization; which implies the main objective function is divided into two different objectives that are optimized alternatively. Precisely, the DR part is optimized while keeping the K-means parameters constant, followed by K-means optimization while keeping the DR part constant. In the third type of method~\cite{dkm_fard2020deep}, in place of K-means, deep K-means is used, which is a continuous variant of K-means as mentioned previously. However, to learn the cluster centers and the data representations jointly, DKM~\cite{dkm_fard2020deep} proposed a continuous variant of K-means, where the argmax of K-means is replaced with one of its soft variants, which is the softmax function. Using deep K-means in place of classical K-means allows to replace the discrete optimization steps with joint optimization AE's parameters and clustering parameters using SGD.

As DKM employs SGD, clustering and network parameter updates are done for every batch of data. However, in classical K-means, the input space remains constant, which is not the case in batch updates as in DKM. To address this, we propose to separately update the network parameters and clustering parameters. The network parameters are updated by jointly optimizing the reconstruction loss and our proposed CenTering (CT) loss function. The CT loss pushes the latent space into being suitable for clustering. Clustering parameters, i.e., the centroids, are then learned on the latent space by optimizing the classical K-means objective. Therefore, the network and the clustering parameter updates are done alternatively for every training epoch. Experiments on some benchmark datasets show that our method can achieve a better Normalized Mutual Information (NMI) and ACCuracy (ACC) score comparatively. In the rest of the paper, we present a brief literature survey in Sec.~\ref{related_works} and our methodology in Sec.~\ref{methodology}. We present qualitative and quantitative experiments with an ablation study in Sec.~\ref{experiments}. Finally, we conclude in Sec.~\ref{conclusion}.

\textbf{Notations}: Before going further, let us introduce the notations.
We consider a dataset of $N$ points, $\mathbf{X} = \{\textbf{x}_i \in \mathbb{R}^{m}: i = 1, \cdots, N\}$, where $m$ is the dimension of the input. In the case of images of shapes ($h \times w \times c$), the data is flattened to the dimension $hwc$, where $h, w, c$ refers to the height, width, and number . of channels of an image. The goal is to cluster $\mathbf{X}$ into $K$ clusters, which is similar in concept to the classes in supervised settings. The set of the centroids of the $K$ clusters is represented by $\mathbf{R} = \{\textbf{r}_1, \textbf{r}_2, \cdots, \textbf{r}_k\}$. Here $\textbf{r}_k$ is the centroid or representative of the cluster $k$. In autoencoder we denote the encoder by $h_{\theta}: \mathbb{R}^m \rightarrow \mathbb{R}^l$, where $l$ is the dimension of latent space or embedding space and $\theta$ is the parameters of the encoder. The decoder is represented by the mapping $h_{\phi}: \mathbb{R}^l \rightarrow \mathbb{R}^m$, where $\phi$ denotes the parameters of the decoder.
We use bold font for vectors. Note that throughout the paper we have used the terms centers and centroids alternatively, while both denote the same thing, i.e., the cluster centers. 


\section{Related Works}
\label{related_works}
The problem of clustering has been well studied over the years. However, studies on leveraging the features learned by deep neural networks for clustering have come to light over the past few years. ~\cite{dec_xie2016unsupervised}~\cite{idec_guo2017improved},~\cite{dcn_yang2017towards},~\cite{dkm_fard2020deep},~\cite{yang2016joint},~\cite{feng2020deep},~\cite{genevay2019differentiable},~\cite{jiang2019dm2c},~\cite{ma2020towards},~\cite{park2021improving} etc.

In this section, we discuss developments in K-means employing deep neural networks. We start by discussing briefly the formulation of the classical K-means~\cite{kmeans_de1994k} algorithm, followed by a brief discussion on the autoencoder (AE), which is used to learn the latent features of the data. Following this, we discuss the formulations of related previous approaches in the context of the current problem.

\subsection{Classical K-means algorithm}
Let $\mathbf{x}$ denotes an object from a set $\{\mathbf{X} = \mathbf{x}_i \in \mathbb{R}^{m}: i = 1, \cdots, N\}$ of objects to be clustered. The task of clustering is to group $N$ data samples into $K$ clusters. The K-means algorithm attempts this task by optimizing the following objective function, 
\begin{equation}
    \begin{split}
    \min_{\mathbf{R}} \sum_{i = 1}^{N} ||\mathbf{x}_i - c(\mathbf{x}_i; \mathbf{R})||_2^2,\\
    \text{with } c(\mathbf{x}_i; \mathbf{R}) = \argmin_{\mathbf{r} \in \mathbb{R}} ||\mathbf{x}_i - \mathbf{r}||_2^2, 
    \end{split}
\end{equation}
where $\mathbf{R} = \{\mathbf{r}_1, \mathbf{r}_2, \cdots, \mathbf{r}_k\}$. Here $\mathbf{r}_k$ is the representative of the cluster k and $\mathbf{R}$ is the set of all the representatives. $c(\mathbf{x}_i; \mathbf{R})$ gives the closest representative of $\mathbf{x}_i$ in terms of L2 distance.

\subsection{Autoencoder}
Among the multiple types of deep neural networks, an autoencoder is a self-supervised deep learner that is trained using an identity function $\mathbf{X} = F_{\theta,\phi}(\mathbf{X}) = g_{\phi}(h_{\theta}(\mathbf{X}))$, where $F$ is the learnable function or the autoencoder as a whole and $h_{\theta}$ and $g_{\phi}$ are respectively the encoder and the decoder function, which represents the mapping from the data space to the latent or encoding space and the reverse mapping from the encoding space back to the data space, respectively.
In general, the objective of AE is, 
\begin{equation}
    \min_{\theta, \phi}^{} \sum_{\mathbf{x} \in \mathbf{X}}|| F_{\theta,\phi}(\mathbf{x}) - \mathbf{x} ||^2_2.
\end{equation}
In general, AE is used for dimensionality reduction (DR) and noise reduction. In problems related to clustering, AE is mostly used for DR to tackle the issue of the curse of dimensionality, which often causes the data space to be unsuitable for clustering.

\subsection{Autoencoder(AE)-based deep clustering variants}
Optimizing the latent space of an AE for clustering was first proposed in DEC~\cite{dec_xie2016unsupervised}. In DEC, Xie et al.~\cite{dec_xie2016unsupervised} proposed an approach that jointly optimizes the centroids $\mathbf{R}$ and the encoder's parameters $\theta$ using Stochastic Gradient Descent (SGD). DEC in particular solves the following objective,
\begin{equation}
    L = \min_{\theta, \mathbf{R}} \text{KL}(P||Q) = \sum_{i} \sum_{j} p_{ij} log \frac{p_{ij}}{q_{ij}},
\end{equation}
where $p_{ij}$ is a function of $q_{ij}$ as the following,
\begin{equation}
    p_{ij} = \frac{q_{ij}^2/\sum_{i}{q_{ij}}}{\sum_{j'}({q_{ij'}^2/\sum_{i}{q_{ij}}})} 
\end{equation}
\begin{equation}
    q_{ij} = \frac{(1+ ||f_{\theta}(\textbf{x}_i) - \textbf{r}_j||_2^2/\alpha^t)^{-\frac{\alpha^t+1}{2}}} {\sum_{j'}{(1+ ||f_{\theta}(\textbf{x}_i) - \textbf{r}_{j'}||_2^2/\alpha^t)^{-\frac{\alpha^t+1}{2}}}}, 
\end{equation}
$q_{ij}$ is the similarity between the embedded point $z_i= f_{\theta}(\textbf{x}_i)$ and the $j^{th}$ cluster centroid $\mathbf{r}_{j}$, interpreted as the probability of assigning sample $i$ to cluster $j$. This is measured using Student's t-distribution as a kernel with $\alpha^t$ as the degrees of freedom.

In IDEC, Guo et al.~\cite{idec_guo2017improved} claimed that employing only the clustering loss might not preserve the local structure of the data in the latent space, leading to a corrupted latent space. Therefore, in~\cite{dcn_yang2017towards} the objective is revised to the following,
\begin{equation}
\min_{\theta, \phi, \mathbf{R}} \sum_{i = 1}^{N}|| F_{\theta,\phi}(\mathbf{x}_i) - \mathbf{x}_i ||^2_2 + \lambda_{idec} \sum_{i} \sum_{j} p_{ij} log \frac{p_{ij}}{q_{ij}},
\end{equation}
where $\lambda_{idec} > 0$ is a coefficient to control the degree of distortion of the latent space for the goal of clustering. A similar idea is proposed by Yang et al.~\cite{dcn_yang2017towards}, but instead of divergence, they used a L2 norm based distance measure. Their objective function is the following,
\begin{equation}
\label{eq: dcn_}
    \min_{\theta, \phi, \mathbf{R}} \sum_{i = 1}^{N}|| F_{\theta,\phi}(\mathbf{x}_i) - \mathbf{x}_i ||^2_2 + \frac{\lambda_{dcn}}{2} || h_{\theta}(\mathbf{x}_i) - \mathbf{M}\mathbf{s}_i ||_2^2,
\end{equation}
where $\mathbf{M}$ is the matrix with its $k^{th}$ column being the $k^{th}$ centroid $\mathbf{r}_k$ and $\mathbf{s}_i$ is the assignment vector of data point i, having only 1 non-zero value in the position of assigned cluster $k$. $\lambda_{dcn} \geq 0$ is a regularization parameter having a similar role as $\lambda_{idec}$. SGD can not be directly applied to jointly optimize $\theta, \phi, \mathbf{M}, \mathbf{S}_i$ together as $\mathbf{s}_i$ is constrained on a discrete set. Therefore, $(\theta, \phi)$ and $(\mathbf{M}, \mathbf{S}_i)$ are optimized in alternating optimization.

To pursue joint optimization DKM~\cite{dkm_fard2020deep} revised the above objective in the following way,
\begin{multline}
\label{eq: dkm_}
    \min_{\theta, \phi, \mathbf{r}_j} \sum_{i = 1}^{N} || F_{\theta,\phi}(\mathbf{x}_i) - \mathbf{x}_i ||^2_2
    + \lambda_{dkm} \sum_{k = 1}^k || h_{\theta}(\mathbf{x}_i) - \mathbf{r}_k ||_2^2 G^{dkm}_{k} (h_{\theta}(\mathbf{x}_i), \alpha_{dkm};\mathbf{R}),
\end{multline}
where, $G^{dkm}_{k}(\cdot)$ is defined as follows,
\begin{equation}
G^{dkm}_{k} (h_{\theta}(\mathbf{x}_i), \alpha_{dkm};\mathbf{R}) = 
\frac{e^{-\alpha_{dkm}|| h_{\theta}(\mathbf{x}_i) - \mathbf{r}_k ||_2^2}}{\sum_{k' = 1}^{K} e^{-\alpha_{dkm} || h_{\theta}(\mathbf{x}_i) - \mathbf{r}_k ||_2^2 }}
\end{equation}
From Eq.\ref{eq: dcn_} to Eq.\ref{eq: dkm_} notice the change is only in the second term where $G^{dkm}_{k}$ is introduced. 

Compared to DKM, our formulation differs in two ways, first, our formulation of $G_{k, f}$ is different; second, we reinitialize the cluster centers after every epoch by applying K-means to the feature space data. IDEC and DCN completely separate feature learning and clustering; instead, we learn clustering-friendly features and then do the clustering alternatively until a convergence criterion based on a loss function is met. Thus, the latent space or feature space is always being learned based on some clustering criterion.

\section{Proposed Formulation}
\label{methodology}
Given a dataset $\mathbf{X}$ having N points $\{\textbf{x}_i \in \mathbb{R}^{m}: i = 1, \cdots, N\}$ and $K$ clusters, our goal is to assign each point to one of the $K$ clusters. We attempt to solve this problem in two steps. In the first step, we learn a feature space that reduces the dimensionality of the data while learning a suitable clustering embedding. For this, we finetune a pretrained autoencoder using two loss functions. First, the reconstruction loss is used to maintain the representability of the data while clustering. Second, our proposed centering (CT) loss minimizes the weighted distance between the cluster centers and the data embeddings. In the second step, we simply optimize the objective of the classical K-means on the data embeddings obtained from the encoder to reinitialize the k cluster centers. Using the above reasoning, we came up with the following objective function,

\begin{equation}
\label{eq: ours}
%
\min_{\theta, \phi, \mathbf{R}}\underbrace{\sum_{i = 1}^{N}(l(g_{\phi}(h_\theta(\mathbf{x}_i)), \mathbf{x}_i)}_{\text{reconstruction loss}} +
\lambda 
\underbrace{\sum_{k=1}^{K} || h_{\theta}(\mathbf{x}_i) - \mathbf{r}_k||^2_2 G_{K,f} (h_{\theta} ({\mathbf{x}_i}), \alpha; \mathbf{R}))}_{\text{centering loss}},
\end{equation}
where,
\begin{equation}
    G_{K,f}(h_{\theta} ({\mathbf{x}_i}), \alpha; \mathbf{R}) = \frac{\frac{1}{f(h_{\theta} ({\mathbf{x}_i}),\mathbf{r}_k)^{\alpha}}}
    {\sum_{k' = 1}^{K} {\frac{1}{f(h_{\theta} ({\mathbf{x}_i}),\mathbf{r}_{k'})^\alpha}}},
\end{equation}
where $f(\cdot, \cdot) = ||\cdot - \cdot||_2^2$. Here, $G_{K, f}(\cdot , \cdot)$ is a differentiable function with respect to $\theta, \mathbf{R}$. $\alpha \in \mathbb{R^+}$ is a parameter. In general we have observed that $\alpha \ge 2$ gives better clustering performance. Eq.~\ref{eq: ours} is optimized using SGD. 

Since our goal is to learn some K-means friendly data representation, therefore, at the end of every epoch we compute K-means on the latent space to initialize the centers $\mathbf{R}$ for the next epoch. Therefore, the values of $\mathbf{R}$ are only used to learn the latent representation of the data during the SGD which is suitable for clustering. 

The steps of our approach are summarized in the Algorithm.~\ref{algo: 1}.

\begin{algorithm}
\caption{The proposed method}
\label{algo: 1}
\begin{algorithmic}[1]
\Require{$A_{1} \dots A_{N}$} 
\Ensure{$Sum$ (sum of values in the array)}

\State Initialise $\theta, \phi, \mathbf{R}$ randomly.
\State Pretrain the AE for $n_p$ number of pretraining epochs.
\State Initialise the centers $\mathbf{R}$ using K-means of the latent representations of the dataset, $L_\mathbf{X} = \{h_{\theta}(\mathbf{x_i}), i = 1, \cdots, N\}$
\For{$epoch \gets 1$ to $n_e$ number of epochs}         \For{$batch \gets 1$ to $n_b$ number of batches}
        \State {Sample minibatch of m samples $\{\mathbf{x_1}, \mathbf{x_2}, \cdots, \mathbf{x_m}\}$ from the dataset}
        \State {Optimize the objective function in Eq.~\ref{eq: ours}}
    \EndFor
    \State Compute the latent representations of the dataset, $L_\mathbf{X} = \{h_{\theta}(\mathbf{x_i}), i = 1, \cdots, N\}$
    \State Optimize the K-means objective function on $L_\mathbf{X}$ for finding the centroids $\mathbf{R}$ of the K clusters.
\EndFor
\State \Return
\end{algorithmic}
\end{algorithm}

\section{Experiments}
\label{experiments}
In this section, we conduct qualitative and quantitative experimental analyses. We compare with some benchmark methods on two standard clustering metrics ACCuracy (ACC) and Normalized Mutual Information (NMI) which are discussed below.
\subsection{Datasets}
We conduct experiments on MNIST, USPS, COIL100, CMU-PIE and RCV1-v2 datasets. RCV1-v2 is text dataset and the rest are image datasets. The details of different datasets are given in Table.~\ref{tab: dataset_details}.
\begin{table}[!h]
\centering
\caption{}
\label{tab: dataset_details}
\begin{tabular}{llllll}
\hline
Dataset      & MNIST & USPS   & COIL100 & CMU-PIE  & RCV1-v2 \\
\hline
\#Samples    & 70000 & 11000   & 7200       & 2856      &    10,000  \\
\#Catagories & 10    & 10      & 100           & 68        &  4  \\

Image size & 28 $\times$ 28 & 16 $\times$16 & 128 $\times$ 128 & 32 $\times$ 32  & \textendash\\

Input dimension & 784 & 256  & 16384  & 1024  & 2000\\
\hline
\end{tabular}
\end{table}

\subsection{Evaluation Metrics}
\textbf{Accuracy}: Considering $c_i$ as the cluster assignment of data $x_i$ and $y_i$
as its ground truth cluster label, the \emph{accuracy (ACC)} of a clustering model is defined as, 
\begin{equation}
ACC = \max_m \frac{\sum_{i = 1}^{N} \textbf{1}\{y_i == m(c_i)\}}{N},
\end{equation}
where $m$ ranges over all possible one-to-one mappings between clusters and labels. N is the total number of data samples in the dataset. Intuitively, this metric finds the best match between the algorithm's cluster assignment and the ground truth assignment. In general, the Hungarian algorithm~\cite{kuhn1955hungarian} is best to compute this mapping.
\\\textbf{Normalized Mutual Information (NMI)}: It is an information theory based similarity measure in clustering. It is bounded in [0,1] and equates to 1 when the ground truth and the predicted clustering are equal. Considering $C$ as the cluster distribution, $Y$ as the ground truth, and $H(\cdot)$ as the entropy, the NMI value is computed as,
\begin{equation}
    NMI = \frac{2H(C, Y)}{H(C) + H(Y)}
\end{equation}
NMI approaches 1 when distribution $C$ is similar to $Y$ and in the opposite case NMI approaches 0.

We compare the ACC and NMI scores with that of others in Tables.~\ref{tab: acc},~\ref{tab: nmi}. 

\subsection{Algorithms in comparison}
\begin{itemize}
    \item KM~\cite{kmeans_de1994k}: The classical K-means clustering approach. 
    
    \item AEKM: Here, before applying K-means, the dimensionality of the data is first reduced using an autoencoder (AE). Therefore, the clustering is done in the latent space or embedding space learned by the autoencoder. 
    
    \item DCN~\cite{dcn_yang2017towards}: It proposes a joint dimensionality reduction and clustering approach to recover a `clustering friendly' latent representation. It employs alternating stochastic optimization to update the clustering parameters (i.e., the cluster centroids) and the network parameters alternatively. Here, the parameters of the AE are initialized by pre-training before employing clustering.
    
    \item DKM~\cite{dkm_fard2020deep} - Similar to DCN, it also proposes an autoencoder based approach for clustering. However, unlike DCN, it proposes a continuous variant of the K-means objective function to jointly achieve dimensionality reduction and clustering using gradient descent. The proposed objective function is fully differentiable with respect to both the clustering and the network parameters. Here also, the AE network is pretrained before adding the clustering loss to the objective.
\end{itemize}

\subsection{Experimental settings}
For every method, we report the average score over 10 runs with different seeds. Note that, same set of 10 seeds is taken for every method to maintain fairness in comparison. For the methods requiring pretraining of the AE model, i.e., DCN, DKM, and ours, we have pretrained for 50 epochs. The finetuning is done for 100 epochs. A fixed batch size of 256 is used for all the experiments. The optimal parameter values for all the parametric methods, i.e., DCN, DKM, and ours, are obtained by grid-search over the feasible set of parameter values. The optimal parameter values are reported in Tab. ~\ref{tab: parameters}.

\subsection{Quantitative analysis}

The ACC and NMI scores reported in Table.~\ref{tab: acc} and Table.~\ref{tab: nmi} show that our method achieves better scores in terms of both metrics. Tab.~\ref{tab: parameters} shows the values of $\lambda$, the coefficient of the clustering loss. $\lambda$ keeps the balance between the reconstruction and the clustering loss to achieve optimal clustering performance. We see that for the CMU-PIE and RCV1-v2 datasets, the $\lambda$ value is low, indicating that the deep embeddings from the pretrained model are relatively good for clustering. Whereas for the MNIST, USPS, and COIL100 datasets, the coefficient of the clustering loss is quite large comparatively, indicating the importance of the clustering loss along with the centroid reinitialization approach for improved clustering outcomes. 
An important observable in Table.~\ref{tab: parameters} is that the $\lambda$ values of ours on the MNIST, USPS, and COIL100 datasets are higher compared to that of DKM, which shows that our approach plays a more significant role in learning clustering-related features compared to that of DKM in these datasets.

 

\begin{table}[!t]
\scriptsize
\setlength{\tabcolsep}{8pt}
\centering
\caption{ACC values over different datasets.}
\label{tab: acc}
\begin{tabular}{cccccc}
\hline
Dataset & MNIST        & USPS    & COIL100        & CMU-PIE        & RCV1         \\
\hline
KM      & 53.50 $\pm$ 0.30   & 67.3 $\pm$ 0.10 & 49.51 $\pm$ 1.13 & 21.31 $\pm$ 0.68 & 50.8 $\pm$ 2.90   \\
AEKM    & 80.80 $\pm$ 1.80   & 72.9 $\pm$ 0.80 & 49.66 $\pm$ 0.84 & 24.17 $\pm$ 1.54 & 56.70 $\pm$ 3.60   \\
DCN     & 81.10 $\pm$ 1.90   & 73.0 $\pm$ 0.80 & 49.23 $\pm$ 0.88 & 24.78 $\pm$ 1.66 & 56.70 $\pm$ 3.60   \\
DKM     & 84.00 $\pm$ 2.20   & 75.7 $\pm$ 1.30 & 49.50 $\pm$ 0.78   & 31.61 $\pm$ 0.86   & 58.3 $\pm$ 3.80   \\
Ours  & \textcolor{blue}{90.01 $\pm$ 5.83} & \textcolor{blue}{78.74 $\pm$ 4.21} & \textcolor{blue}{51.56 $\pm$ 0.94}   & \textcolor{blue}{32.40 $\pm$ 1.81}   & \textcolor{blue}{60.39 $\pm$ 2.33}\\


\hline
\end{tabular}
\end{table}
\begin{table}[!t]
\scriptsize
\setlength{\tabcolsep}{8pt}
\centering
\caption{NMI values over different datasets.}
\label{tab: nmi}
\begin{tabular}{cccccc}
\hline
Dataset & MNIST        & USPS                  & COIL100        & CMU-PIE        & RCV1         \\
\hline
KM   & 49.8 $\pm$ 0.5   & 61.4 $\pm$ 0.1   & 76.82 $\pm$ 0.35 & 41.67 $\pm$ 0.67 & 31.3 $\pm$ 5.4   \\
AEKM & 75.2 $\pm$ 1.1   & 16.9 $\pm$ 1.3   & 77.03 $\pm$ 0.30 & 51.43 $\pm$ 1.87 & 31.5 $\pm$ 4.3   \\
DCN  & 75.7 $\pm$ 1.1   & 71.9 $\pm$ 1.2   & 76.76 $\pm$ 0.47 & 52.33 $\pm$ 1.88 & 31.6 $\pm$ 4.3   \\

DKM  & 79.6 $\pm$ 0.9   & 77.6 $\pm$ 1.1   & 77.82 $\pm$ 0.25   & 61.92 $\pm$ 0.82   & 33.1 $\pm$ 4.9 \\
Ours & \textcolor{blue}{87.64 $\pm$ 1.55} & \textcolor{blue}{80.64 $\pm$ 1.20} & \textcolor{blue}{78.17 $\pm$ 0.40}   & \textcolor{blue}{62.18 $\pm$ 0.89}   & \textcolor{blue}{35.85 $\pm$ 3.03}\\



\hline
\end{tabular}
\end{table}

\begin{table}
\centering
\caption{Optimal values of parameters of different methods.}
\label{tab: parameters}
\begin{tabular}{c|cccccccc}
\hline
Dataset & \begin{tabular}[c]{@{}c@{}}Pretraining\\ epochs\end{tabular} & \begin{tabular}[c]{@{}c@{}}Fine-tuning\\ epochs\end{tabular} & Batch size  & $\alpha$ & $\lambda_{\text{Ours}}$ & $\lambda_{\text{DKM}}$ & $\lambda_{\text{DCN}}$\\
\hline
MNIST & 50 & 100 & 256 & 3 & 1e+1 & 1e+0 & 1e+1\\
USPS & 50 & 100 & 256 & 2.5 & 1e+1 & 1e+0 & 1e-1 \\
COIL100 & 50 & 100 & 256 & 3 & 1e+1 & 1e-1 & 1e+0\\
CMU-PIE & 50 & 100 & 256 & 2 &1e-3 & 1e-2 & 1e-1\\
RCV1 & 50 & 100 & 256 & 2 & 1e-4 & 1e-2  & 1e-1\\

\hline
\end{tabular}
\end{table}


\subsection{Qualitative Analysis}
\label{qualitative}
\begin{figure}[!htp]
\centering
		\includegraphics[width=0.8\linewidth]{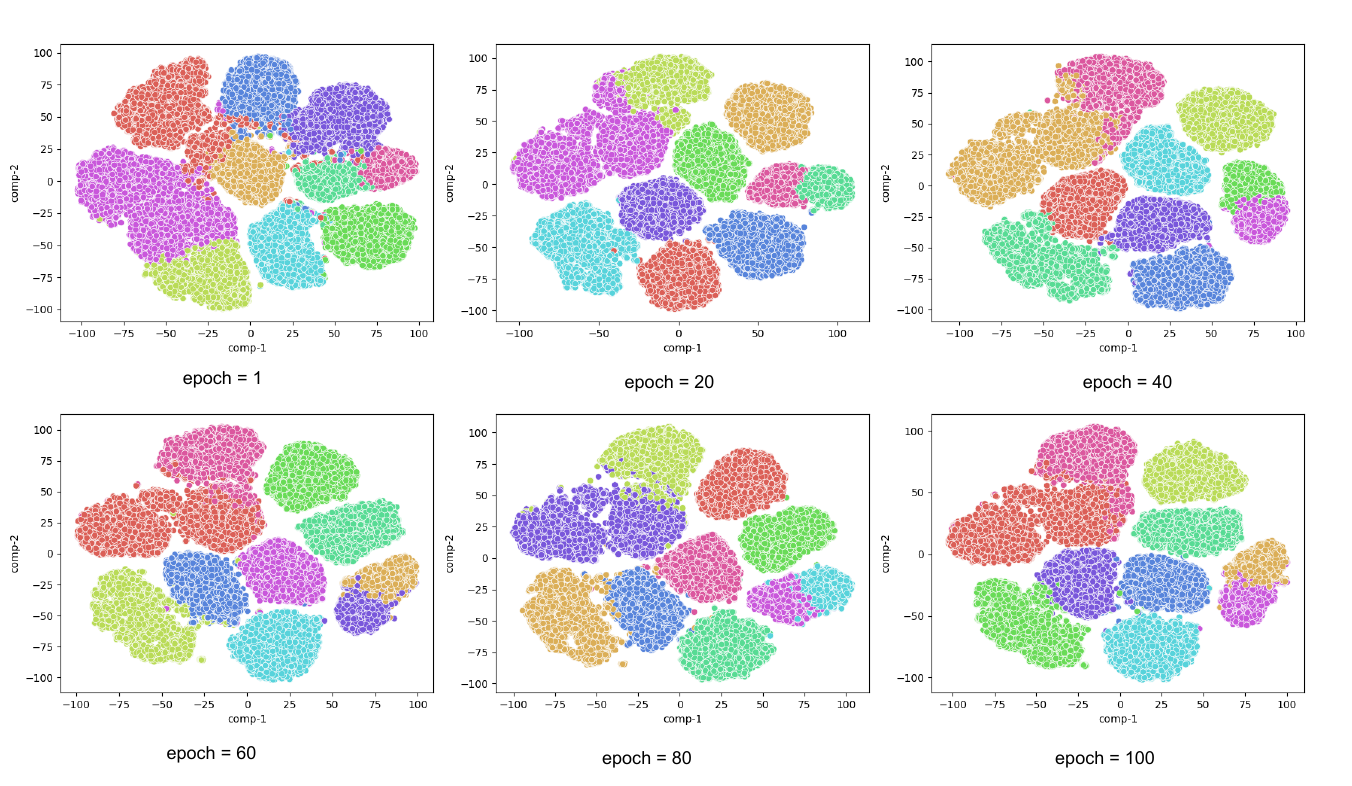}
		\caption{figure}{The visualization of the learned clusters in the latent space of the full MNIST dataset over the different fine-tuning epochs of our method.}
		\label{fig: MNIST_pretrain_vis1_ours}
\end{figure}

\begin{figure}[!ht]
		\centering
		\includegraphics[width=0.8\linewidth]{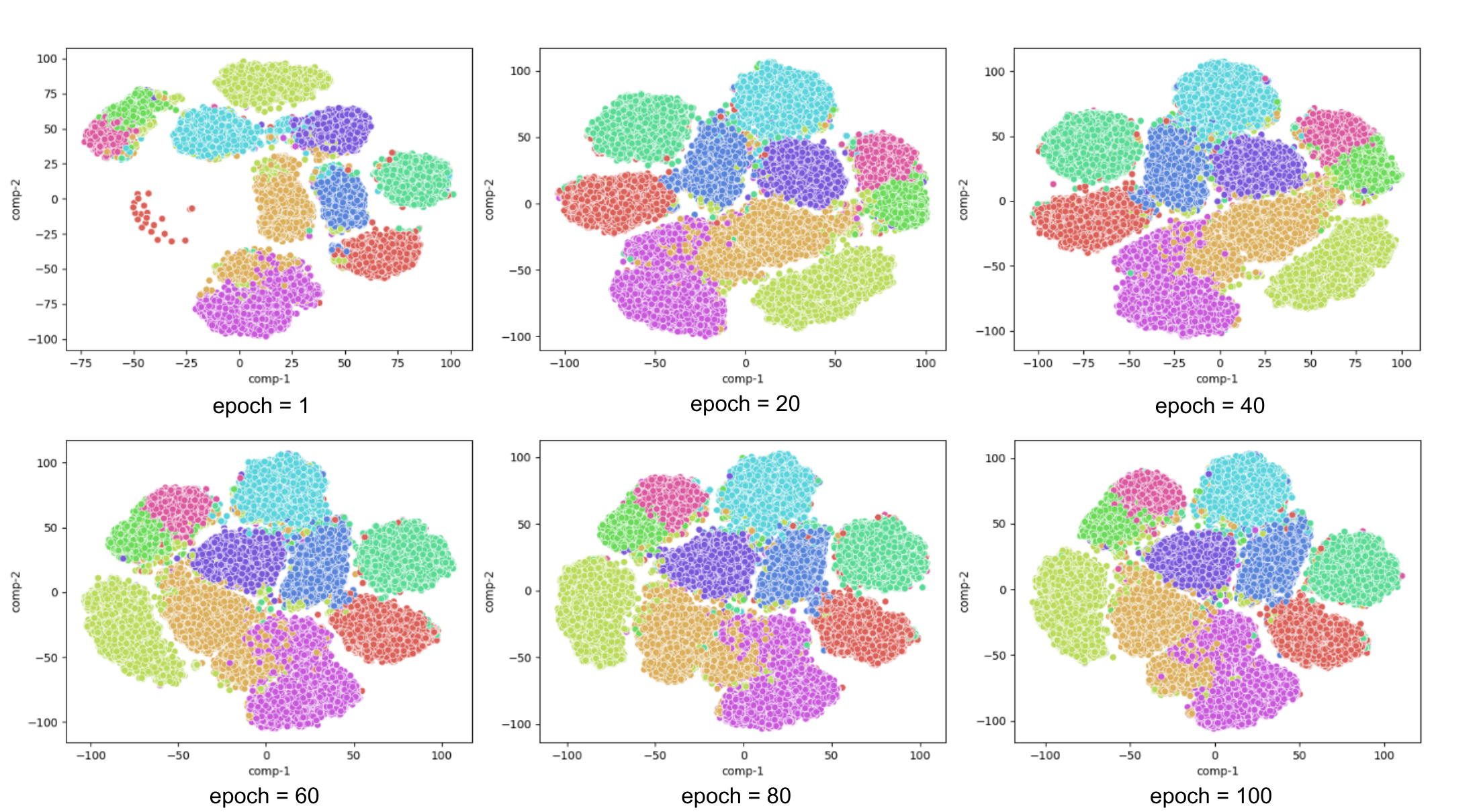}
		\caption{figure}{The visualization of the learned clusters in the latent space of the full MNIST dataset over different fine-tuning epochs of DKM.}
		\label{fig: MNIST_pretrain_vis1_dkm}
\end{figure}

To compare with DKM qualitatively, we have given the TSNE~\cite{tsne} plot of deep embeddings of the full-MNIST dataset with the different predicted cluster labels of our method and that of DKM in Fig.~\ref{fig: MNIST_pretrain_vis1_ours},~\ref{fig: MNIST_pretrain_vis1_dkm}, respectively. In order to analyze the clustering progression, we provided the plot over the different fine-tuning epochs. For fair comparison, both methods have been executed under the same experimental conditions.
It can be observed that from the first fine-tuning epoch, our method shows better cluster compactness that improves as the epochs increase. Finally, we see that our clustering shows increased inter-cluster distance compared to that of DKM, resulting in better clustering metric values; that can be verified from the quantitative analysis.
\begin{figure}[!h]
		\centering
		\includegraphics[width=0.7\linewidth]
{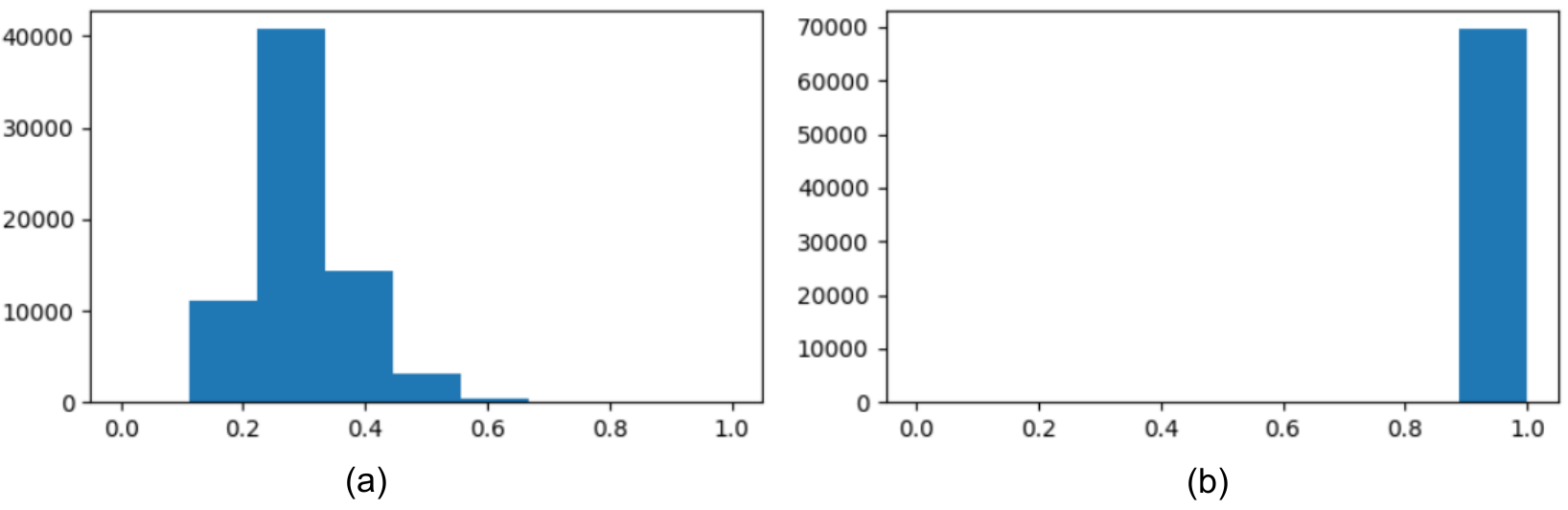}
		\caption{figure}{Histograms of the highest cluster membership value of each data point on MNIST, (a)~before finetuning, (b)~after finetuning at the 100$^{th}$ epoch. Observe that before finetuning the cluster memberships are in the low confidence region, indicating poor clustering performance. Whereas after finetuning all the memberships are in high confidence, indicating better clustering.}
		\label{fig: MNIST_hist_ours}
\end{figure}
\begin{figure}[!h]
		\centering
		\includegraphics[width=0.5\linewidth]{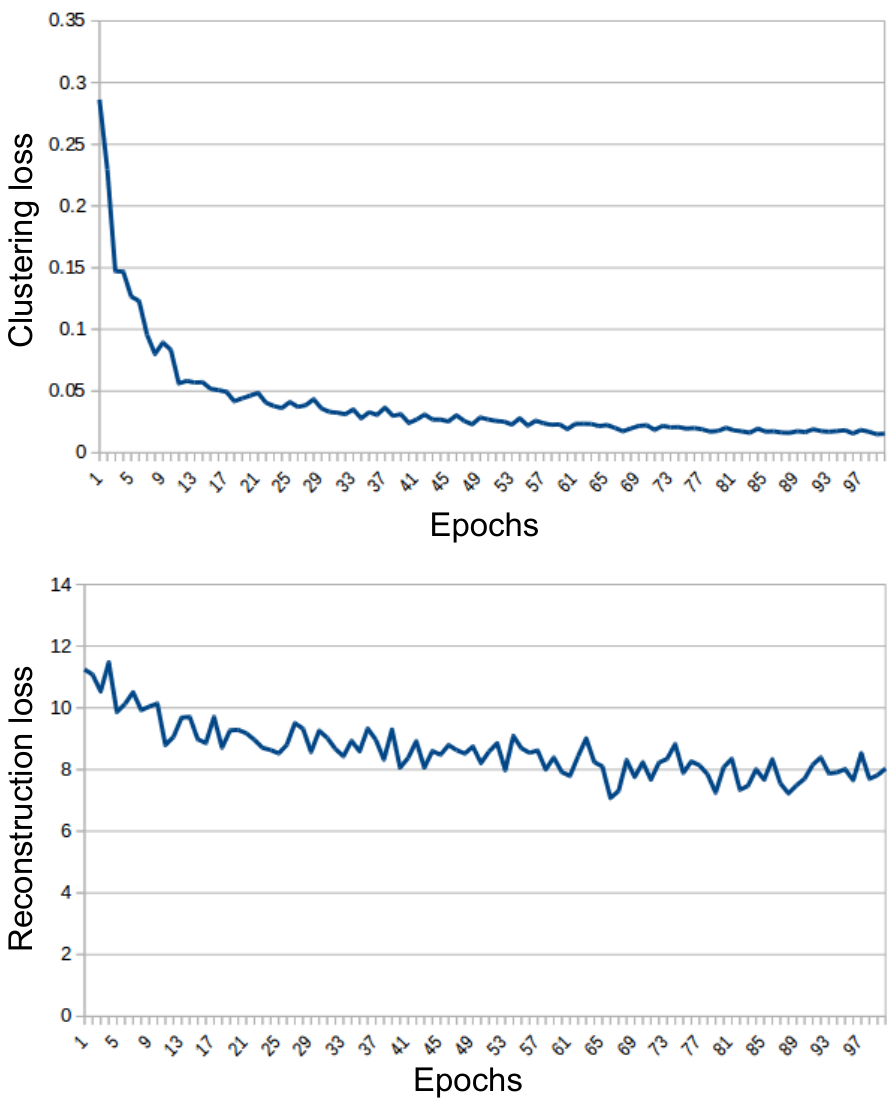}
		\caption{figure}{
		Loss values over different epochs in the finetuning stage on MNIST dataset. Observe that the clustering loss decreases more compared to the reconstruction loss. This is because the reconstruction is optimized mostly in the pretraining stage. During the finetuning stage, the clustering loss plays a major role, while the reconstruction loss is mainly to keep the data representability unaffected in pursuing the goal of clustering.}
		\label{fig: MNIST_loss_curves_ours}
\end{figure}
We also present some graph visualizations to showcase the distribution of the points across various clusters over different epochs. From Fig.~\ref{fig: MNIST_hist_ours}, we see that after pre-training, the membership values are in the low confidence region. However, as the learning progresses, the memberships are mostly in the high confidence region, indicating the learning of the clustering parameters. In Fig.~\ref{fig: MNIST_loss_curves_ours}, we present the plot of the clustering and reconstruction losses of our method over the 100 fine-tuning epochs on the MNIST dataset. Observe that the losses decrease over time, showing progress towards convergence.
\subsection{Ablation Study}
In this section, we conduct an ablation study to understand the significance of the centroid reinitialization (say, rein) strategy in the proposed approach. While our method shows improvement over the compared methods on the presented datasets, to understand the effect of the centroid-rein strategy, we show the results of an instance of our approach without the rein-strategy (denoted as Ours$^{-rein}$). The results in terms of both the metrics ACC and NMI are presented in Tab.~\ref{tab: ablation_acc} and~\ref{tab: ablation_nmi}. It can be observed that our method without the rein strategy does not improve over the compared methods. Which implies that the centroid rein strategy is crucial to the success of the proposed approach. However, if we join the centroid-rein strategy with the approach of DKM (denoted as DKM$^{+rein}$), no improvement is observed, which can be verified from the results presented in Tab.~\ref{tab: ablation_acc} and~\ref{tab: ablation_nmi}, where the results of DKM$^{+rein}$ are in fact inferior to those of DKM. This shows that the usefulness of the rein strategy depends on the choice of the clustering specific loss.

Please note that, in DKM$^{+rein}$, we reinitialize the cluster centroids after every finetuning epoch of the clustering phase of DKM. We edited the published code of DKM~\footnote{https://github.com/MaziarMF/deep-K-means} to get this result.
%
%
\begin{table}[!htp]
\scriptsize
\setlength{\tabcolsep}{8pt}
\centering
\caption{Ablation study: understanding the role of centroid reinitialization strategy in terms of the ACC metric. Method$^{+rein}$ and Method$^{-rein}$ denote the method with and without reinitialization strategy, respectively.}

\label{tab: ablation_acc}

\begin{tabular}{cccccc}
\hline
Dataset & MNIST        & USPS    & COIL100        & CMU-PIE        & RCV1         \\
\hline
DCN     & 81.10 $\pm$ 1.90   & 73.0 $\pm$ 0.80 & 49.23 $\pm$ 0.88 & 24.78 $\pm$ 1.66 & 56.70 $\pm$ 3.60   \\
DKM     & 84.00 $\pm$ 2.20   & 75.7 $\pm$ 1.30 & 49.50 $\pm$ 0.78   & 31.61 $\pm$ 0.86   & 58.3 $\pm$ 3.80   \\

DKM$^{+rein}$  & 58.30 $\pm$ 4.22 & 52.66 $\pm$ 2.59 & 28.27 $\pm$ 0.94 & 23.70 $\pm$ 5.89 &  45.52 $\pm$ 2.99\\

Ours$^{-rein}$ & 85.49 $\pm$ 4.92 & 70.12 $\pm$ 1.93  & 42.71 $\pm$ 6.36 & 15.39 $\pm$ 1.74 & 58.55 $\pm$ 0.23\\

Ours  & \textcolor{blue}{90.01 $\pm$ 5.83} & \textcolor{blue}{78.74 $\pm$ 4.21} & \textcolor{blue}{51.56 $\pm$ 0.94}   & \textcolor{blue}{32.40 $\pm$ 1.81}   & \textcolor{blue}{60.39 $\pm$ 2.33}\\

\hline
\end{tabular}
\end{table}
\begin{table}[!htp]
\scriptsize
\setlength{\tabcolsep}{8pt}
\centering
\caption{Ablation study: understanding the role of centroid reinitialization strategy in terms of the NMI metric. Method$^{+rein}$ and Method$^{-rein}$ denote the method with and without re-initialization strategy, respectively.}
\label{tab: ablation_nmi}
\begin{tabular}{cccccc}
\hline
Dataset & MNIST        & USPS                  & COIL100        & CMU-PIE        & RCV1         \\
\hline
DCN  & 75.7 $\pm$ 1.1   & 71.9 $\pm$ 1.2   & 76.76 $\pm$ 0.47 & 52.33 $\pm$ 1.88 & 31.6 $\pm$ 4.3   \\
DKM  & 79.6 $\pm$ 0.9   & 77.6 $\pm$ 1.1   & 77.82 $\pm$ 0.25   & 61.92 $\pm$ 0.82   & 33.1 $\pm$ 4.9   \\

DKM$^{+rein}$  & 49.43 $\pm$ 3.35 & 46.41 $\pm$ 2.20 & 55.57 $\pm$ 0.77 & 44.14 $\pm$ 9.61 & 16.60 $\pm$ 2.35\\


Ours$^{-rein}$  & 86.88 $\pm$ 1.36 & 76.64 $\pm$ 0.82 & 72.36 $\pm$ 3.06 & 46.51 $\pm$ 2.17 & 33.18 $\pm$ 0.25 \\

Ours & \textcolor{blue}{87.64 $\pm$ 1.55} & \textcolor{blue}{80.64 $\pm$ 1.20} & \textcolor{blue}{78.17 $\pm$ 0.40}   & \textcolor{blue}{62.18 $\pm$ 0.89}   & \textcolor{blue}{35.85 $\pm$ 3.03}\\
\hline
\end{tabular}
\end{table}
%


\section{Conclusions and Future works}
\label{conclusion}
This paper introduces a centroid-based clustering method that improves on the existing deep neural network-based K-means approaches. Along with our proposed clustering-specific loss function, we proposed the idea of employing centroid reinitialization after every fintuning epoch in the clustering phase. We have empirically shown the importance of this reinitialization. While the idea presented in this paper is verified empirically, a thorough theoretical justification is required in the future. An in-depth comparative analysis of the difference between our formulation and that of our closest deep clustering variant is worth doing in the future to gain better clarity. In the future, we also plan to extend this idea towards improving the performance of the without pre-training case. Instead of initializing the cluster centers with K-means, how a random centroid initialization can achieve similar performance with pre-training can also be a good line of research in the future. 

\bibliographystyle{plain}
\bibliography{references}
\end{document}